\documentclass{nature}

\usepackage{amssymb, amsmath}
\usepackage[hidelinks]{hyperref}
\usepackage{graphicx}
\usepackage{capt-of}
\usepackage{booktabs}

\usepackage{algorithm}
\usepackage{algpseudocode}

\usepackage{lineno}

\usepackage{tabularx}
\newcommand\setrow[1]{\gdef\rowmac{#1}#1\ignorespaces}
\newcommand\clearrow{\global\let\rowmac\relax}
\clearrow

\title{Neuromorphic Computing for\\ Content-based Image Retrieval}

\author{Te-Yuan Liu$^{1}$, Ata Mahjoubfar$^{1,*}$, Daniel Prusinski$^{1}$, \& Luis Stevens$^{1}$}

\begin{document}

\maketitle

\begin{affiliations}
\item Target Tech, 100 S. Mathilda Place, Suite 300, Sunnyvale, CA 94086.
\item[$^{*}$] Corresponding author: \href{mailto:ata.mahjoubfar@target.com}{ata.mahjoubfar@target.com}
\end{affiliations}

\section*{Abstract}
\begin{linenumbers}
Neuromorphic computing mimics the neural activity of the brain through emulating spiking neural networks. In numerous machine learning tasks, neuromorphic chips are expected to provide superior solutions in terms of cost and power efficiency. Here, we explore the application of Loihi, a neuromorphic computing chip developed by Intel, for the computer vision task of image retrieval. We evaluated the functionalities and the performance metrics that are critical in content-based visual search and recommender systems using deep-learning embeddings. Our results show that the neuromorphic solution is about 2.5 times more energy-efficient compared with an ARM Cortex-A72 CPU and 12.5 times more energy-efficient compared with NVIDIA T4 GPU for inference by a lightweight convolutional neural network without batching while maintaining the same level of matching accuracy. The study validates the potential of neuromorphic computing in low-power image retrieval, as a complementary paradigm to the existing von Neumann architectures.
\end{linenumbers}

\section*{Introduction}
\begin{linenumbers}
Neuromorphic computing is a non-von Neumann computer architecture, aiming to obtain ultra-high-efficiency machines for a diverse set of information processing tasks by mimicking the temporal neural activity of the brain\cite{james2017historical,wunderlich2019demonstrating,cauwenberghs1998neuromorphic}. In neuromorphic computing, numerous spiking signals carry information among computing units i.e. artificial neurons, synchronously or asynchronously\cite{schuman2017survey}, forming a mesh-like, nonlinear dynamical system\cite{neckar2018braindrop}. The information can be encoded in the temporal characteristics of the signals, for example firing rates\cite{ponulak2011introduction}.

In this work, we implement and analyze a low-latency computer vision model for visual search engines and recommender systems that evaluate the visual similarity between a query image and a database of product images. In conventional machine learning pipelines, this is often performed by transfer learning using a deep convolutional neural network (CNN)\cite{lecun2015deep} pre-trained on a large-scale dataset e.g., ImageNet\cite{deng2009imagenet,kornblith2019better} and fine-tuned on a domain-specific image dataset e.g., DeepFashion2 for apparel\cite{ge2019deepfashion2}. The embeddings of the images are calculated by inferring the activation values of the last few layers of the neural network as visual features\cite{babenko2014neural,chen2016deep,gordo2016deep,mahjoubfar2017time,gordo2017end,li2019deep,noh2017large}. The distances between embeddings of the query image and the database images are used to find the nearest neighbors for the query image in the embeddings space, identifying the most similar items visually\cite{cao2017deep}.

Here, we evaluate the same visual search and recommendation technique using embeddings generated by the neuromorphic neural networks. We train spiking convolutional neural networks on a clothing-specific image classification dataset, Fashion-MNIST\cite{xiao2017fashion}. The trained spiking neural networks are then used for extraction of features for the product images and the query images. The embeddings will be based on the patterns of the temporal spikes, and similar to the conventional convolutional neural networks, they are used for finding nearest visual neighbors of the query image among product images. Our results show considerable power efficiency in finding the most visually similar products using neuromorphic chips and particularly Loihi\cite{davies2018loihi}.
\end{linenumbers}

\section*{Methods}
\begin{linenumbers}
To explore applications of neuromorphic computing in image retrieval, we built and deployed a spiking neural network (SNN) on Intel's Loihi neuromorphic chip. Our image search pipeline is shown in Fig.\ref{pipeline}. Firstly, we convert a trained artificial neural network (ANN) into a spiking neural network (SNN) and deploy it on Loihi chip. We then feed training and test images into the SNN and probe the neurons of the layer before the output layer to get image embeddings. Finally, nearest neighbor search is employed on CPU cores to find the best matches in the training dataset for each test image.

\begin{center}
\includegraphics[scale=1]{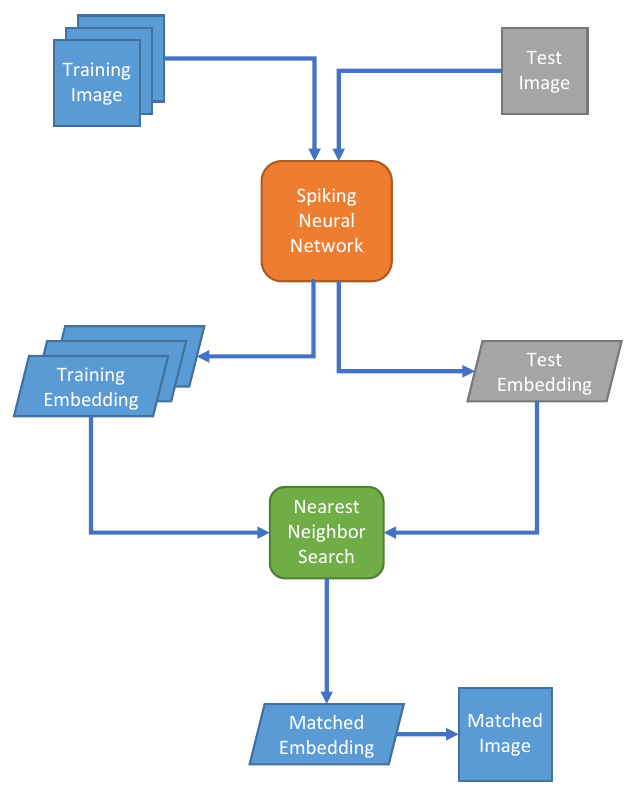}
\captionof{figure}[caption]{
\label{pipeline}
Pipeline of image retrieval by spiking neural network.
} 
\end{center}

In the first step, we train different ANNs by minimizing the cross-entropy loss function for the classification of the Fashion-MNIST dataset via backpropagation. Then, we convert the ANNs into SNNs, compare the classification test accuracies of SNNs, and select the most accurate SNN model. Suggested by Hunsberger and Eliasmith\cite{hunsberger2016training} and Sengupta et al.\cite{sengupta2019going}, we reduce the feature map size using average pooling rather than max pooling, and employ dropout to regularize\cite{srivastava2014dropout}.

Note that there are two constraints on the neural network architectures that can be deployed on Loihi chips. One constraint is that the synaptic memory, which stores neuron weights per neuromorphic core is 128 KB. This indicates that the number of parameters associated with neurons in a core is limited. The other constraint is the maximum fan-in of 4,096 per neuromorphic core, which means the input size of the neurons cannot exceed 4,096\cite{rajendran2019low}. These two constraints result in neural networks deployed on Loihi chip to have relatively slim layers rather than wide layers.

Given an ANN, the conversion is done through building a SNN which has the same architecture as the ANN, but changing the neuron type to Leaky, Integrate and Fire (LIF) neuron with soft-reset, which is a variant of Residual Membrane Potential (RMP) neuron proposed by Han et al.\cite{han2020rmp}. Then, floating-point ANN parameters are scaled to integers and transplanted to the SNN as Loihi chip executes operations with integer numbers. The spiking threshold of each LIF neuron is determined at the same time as the parameter scaling, using a method provided by Loihi NxSDK\cite{lin2018mapping}. The method of parameter scaling and threshold calculation is shown in Algorithm \ref{alg} (For more details, see mapping spiking neural networks onto a manycore neuromorphic architecture by Lin et al.\cite{lin2018mapping}).

\begin{algorithm}
\caption{Parameter scaling and threshold calculation}
\label{alg}
	\begin{algorithmic}[1]
	\Require Normalized Input: $input\in [0, 1]^{N\times H\times W\times C}$
	\State $W_{MAX} = 2^{num\_weight\_bits - 1} - 1, b_{MAX} = 2^{num\_bias\_bits - 1} - 1$
	\State $slope = 1, param\_percentile=99.999, activation\_percentile=99.999$
	\For{$snn\_layer$, $ann\_layer$ in zip($SNN.layers$, $ANN.layers$)}
		\If{$snn\_layer$ is input layer}
			\State $param\_scale = W_{MAX}$
			\State $dvdt = input\times param\_scale$
		\Else
			\State $weight, bias = ann\_layer.get\_param()$
			\State $bias = bias\times slope$
			\State $weight\_norm = percentile(abs(weight), param\_percentile)$
			\State $bias\_norm = percentile(abs(bias), param\_percentile)$
			\State $weight\_ratio = \frac{W_{MAX}} {weight\_norm}, bias\_ratio = \frac{b_{MAX}} {bias\_norm}$
			\State $param\_scale = min(weight\_ratio, bias\_ratio)$
			\State $weight = int(weight\times param\_scale), bias = int(bias\times param\_scale)$
			\State $snn\_layer.set\_param(weight, bias)$
			\State $dvdt = ReLU(weight\cdot spikerate + bias)$
		\EndIf
		\State $threshold = int(percentile(dvdt, activation\_percentile))$
		\State $snn\_layer.threshold = threshold$
		\State $spikerate = min(\frac{dvdt} {threshold}, 1)$
		\State $slope = slope\times \frac{param\_scale} {threshold}$
	\EndFor
	\end{algorithmic}
\end{algorithm}

Similar to the spike-norm algorithm proposed by Sengupta\cite{sengupta2019going}, a set of images are fed into the network and the threshold at each layer is set to the maximum activation at that layer. However, Loihi chip uses a rate-based simulation of SNN instead of doing the actual SNN forward-pass to calculate the spiking thresholds.

In Algorithm \ref{alg}, there are two important variables. One is named $param\_scale$, which gives the factor we use to scale the ANN parameters to integers to get the SNN parameters. The other one is named $threshold$, which is the spiking threshold that decides the LIF neuron spiking activity.

Algorithm \ref{alg} requires a batch of input images to tune the spiking threshold, and they are represented as an $N\times H\times W\times C$ matrix with floating-point elements ranging between zero and one; $N$ for the number of images and $H$, $W$, $C$ for the image's height, width, and channel. Line 1 set the $W_{MAX}$, $b_{MAX}$, and line 2 set the $slope$, $param\_percentile$, $activation\_percentile$ variables. If we use 9 bits to represent SNN weights on Loihi chip, then the maximum weight $W_{MAX}$ is $2^{9 - 1} - 1 = 255$. We set the maximum bias $b_{MAX}$ in the same way. The $slope$ variable shows the ratio between the SNN neuron output and the ANN neuron output at the current layer and is initialized to one.

In line 3, we get $snn\_layer$ and its corresponding $ann\_layer$. From line 4 to 6, if $snn\_layer$ is the input layer which encodes input images into spike time series, we set $param\_scale$ to $W_{MAX}$ and multiply $input$ by $param\_scale$ to get the $dvdt$, which is the neuron membrane potential increment rate. Note that $dvdt$ here still has the shape of $N\times H\times W\times C$ as the input layer only multiplies the input by a scalar. 

From line 7 to 16, if the $snn\_layer$ is not the input layer, we have to scale the ANN parameters and set the SNN parameters. In line 8, we get the ANN $weight$ and $bias$ from $ann\_layer$. Then in line 9, we multiply $bias$ by $slope$ to update $bias$ with the scaling of the previous layer. In line 10, we set $weight\_norm$ as one single value by getting a percentile value of $abs(weight)$ and do likewise to set $bias\_norm$ in line 11. Then in line 12, we set $weight\_ratio$ as the ratio between $W_{MAX}$ and $weight\_norm$ to find out how many times we can scale up $weight$ without exceeding $W_{MAX}$, and we do the same thing to calculate $bias\_ratio$. In line 13, we compare $weight\_ratio$ and $bias\_ratio$ to set the $param\_scale$ to the smaller value. In line 14 and 15, we use $param\_scale$ to scale the ANN $weight$ and $bias $, quantizing them to integers, and set them as the parameters of $snn\_layer$. In line 16, we calculate $dvdt$ by simulating the ANN neuron activation and the shape of $dvdt$ becomes $N\times FH\times FW\times FC$, where $FH$, $FW$, and $FC$ stand for the feature map's height, width, and channel.

In line 18 and 19, we set the threshold of neurons at $snn\_layer$ to the quantized percentile value of $dvdt$ so there is one single threshold value for this layer. Then, in line 20, we calculate the $spikerate$, an estimation of the spiking probability of neurons, as the output of $snn\_layer$, which has the same shape as $dvdt$. In line 21, we update $slope$ by multiplying it with the ratio of $param\_scale$ and $threshold$.

Now having a SNN at hand, we start feeding images into the network. For each image, we probe the neurons of the layer before the output layer at the last execution time step to get the neuron membrane potentials. The membrane potential vector is then the embedding of the input image.

Our SNN takes images in the training and test sets as inputs and generates their embeddings. We see the training image embeddings as a corpus of image features. For each test image, we apply nearest neighbor search using cosine similarity to find images in the corpus that are the closest to the test image in the embedding space.
\end{linenumbers}

\section*{Results}
\begin{linenumbers}
We implemented and tested 3-layer, 4-layer, and 5-layer SNNs for classification of Fashion-MNIST dataset. We selected Fashion-MNIST as our evaluation dataset because it is suitable for benchmarking small-footprint computer vision models. Note that we use this dataset without data augmentation in our experiments. The architectures analyzed are shown in Table \ref{architectures_performance_comparison}. In the architecture column, number of convolutional kernels (number of output channels) in each layer are concatenated by hyphens. Note that the last architecture in Table \ref{architectures_performance_comparison} was not deployable on the Loihi chip because the maximum fan-in was exceeded. The fourth architecture in Table \ref{architectures_performance_comparison} scores the best classification test accuracy when converted to a SNN; this architecture is shown in Fig. \ref{snn_architecture}. It consists of three layers, including two convolutional layers and one dense layer. We use this SNN architecture to conduct the rest of the experiments. The image embeddings are generated by flattening the output of the last convolutional layer from a $4 \times 4 \times 64$ tensor to a $1024$-dimensional vector.

\begin{table}
\centering
\caption{Various CNN architectures and their performance.}
\begin{tabular}[t]{>{\rowmac}c|>{\rowmac}c|>{\rowmac}c|>{\rowmac}c|>{\rowmac}c<{\clearrow}}
\toprule
 & ANN & ANN & ANN & SNN \\
Architecture & train & validation & test & test \\
 & acc. (\%) & acc. (\%) & acc. (\%) & acc. (\%) \\
\midrule
4-8-10 & 85.03 & 85.37 & 83.96 & 83.73 \\
8-16-10 & 89.11 & 88.43 & 87.41 & 86.98 \\
16-32-10 & 89.41 & 88.72 & 87.85 & 87.41 \\
\setrow{\bfseries}32-64-10 & 93.99 & 90.68 & 90.07 & 90.01 \\

4-8-16-10 & 87.47 & 87.20 & 86.16 & 86.02 \\
8-16-32-10 & 90.43 & 89.03 & 88.40 & 76.54 \\
16-32-64-10 & 90.36 & 89.25 & 87.91 & 87.85 \\
32-64-128-10 & 93.37 & 90.08 & 89.43 & 88.18 \\

4-8-16-32-10 & 89.14 & 88.57 & 87.87 & 87.49 \\
8-16-32-64-10 & 92.11 & 90.20 & 89.51 & 84.19 \\
16-32-64-128-10 & 94.27 & 90.60 & 89.76 & 85.79 \\
32-64-128-256-10 & 93.38 & 90.78 & 90.21 & N/A \\

\bottomrule
\end{tabular}
\label{architectures_performance_comparison}
\end{table}

\begin{center}
\includegraphics[scale=0.55]{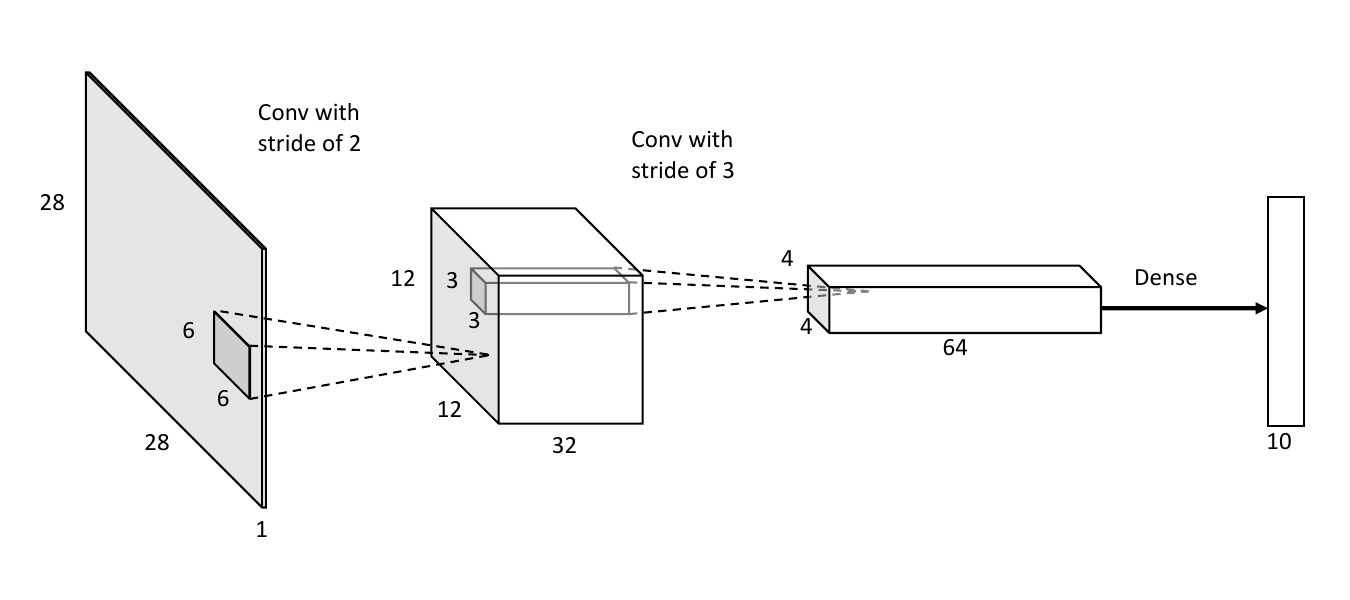}
\captionof{figure}[caption]{
\label{snn_architecture}
Architecture of our best SNN.
} 
\end{center}

The SNN layer partition on a Loihi chip is shown in Fig. \ref{chip0_core_assignment}. There are 128 neuromorphic cores on a Loihi chip in 8 rows with 16 cores a row. We can see each layer occupies certain number of the neuromorphic cores. Our best performing SNN (Fig. \ref{snn_architecture}) is relatively compact, so the number of cores occupied is small compared with the number of cores available on a Loihi chip.

\begin{center}
\includegraphics[scale=0.8]{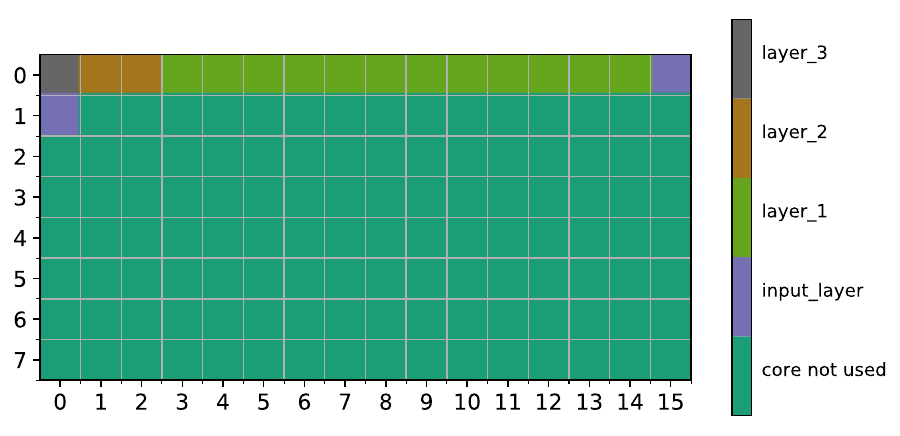}
\captionof{figure}[caption]{
\label{chip0_core_assignment}
SNN layer distribution on Loihi chip.
} 
\end{center}

SNN has an intrinsic execution time parameter, called number of time steps, which is used to define how many discrete time slots are given to the network to process information during inference. It is intuitive that the more time steps we give our SNN to process the information, the higher performance we get, but the runtime is also larger. This tradeoff between performance and number of time steps is shown in  Fig. \ref{performance}. We can see that performance metrics skyrocket between 4 time steps and 16 time steps and then plateau, showing that using 16 time steps is enough to achieve certain degree of performance. The error bars indicate the negligible variations among five independently trained networks, displaying reproducibility of our results.

\begin{center}
\includegraphics[scale=1]{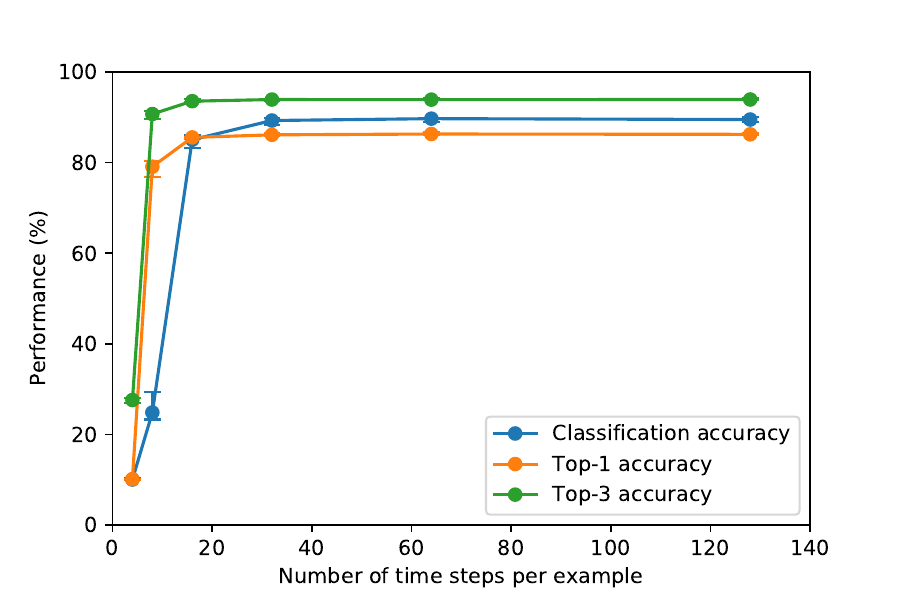}
\captionof{figure}[caption]{
\label{performance}
Tradeoff between performance metrics and number of time steps.
} 
\end{center}

The relation between the runtime and the number of time steps is shown in Fig. \ref{runtime}. As we gradually increase the number of time steps, the runtime scales up almost linearly. However, the runtime is independent of the number of time steps for small numbers, e.g., 4 or 8 time steps because the overhead takes up the majority of the runtime.

\begin{center}
\includegraphics[scale=1]{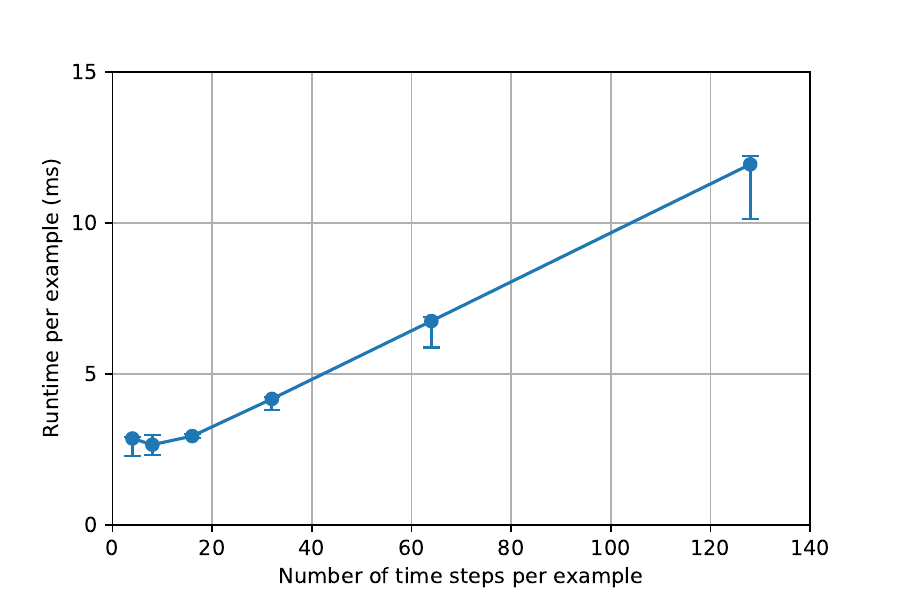}
\captionof{figure}[caption]{
\label{runtime}
Tradeoff between runtime and number of time steps.
} 
\end{center}

The performance comparison between the selected SNN and its ANN counterpart is shown in Table \ref{performance_comparison}. Note that the number in the parentheses next to the model type is the number of time steps used per example during SNN inference. The ANN and SNN have the same network architecture but different neuron types and parameters. We can see that the SNN using 128 time steps have accuracies very close to the ANN, indicating that the SNN is capable of achieving comparable performance with its ANN counterpart. Using fewer time steps, e.g., 16 time steps, our SNN suffers a classification accuracy degradation, but the gap is smaller than 5\%. However, the top-1 and top-3 accuracies of the SNN with 16 time steps is still very close to the ANN. This means that the SNN with 16 time steps per inference generates reasonable embeddings, suitable for the image retrieval task.

\begin{table}
\centering
\caption{Test accuracies.}
\begin{tabular}[t]{cccc}
\toprule
Model Type & Classification & Top-1 & Top-3  \\
 & Accuracy (\%) & Accuracy (\%) & Accuracy (\%) \\
\midrule
ANN & 90.07 & 87.49 & 94.55 \\
SNN (16)  & 85.05 & 85.55 & 93.56 \\
SNN (128) & 90.01 & 86.58 & 93.93 \\
\bottomrule
\end{tabular}
\label{performance_comparison}
\end{table}

Several examples of the SNN image retrieval are shown in Fig. \ref{image_search}. The first column shows ten query images, each from a class in the dataset. The next three columns present three randomly-selected images from the corpus with the same class label as the query images. The next three columns demonstrate the top-three images selected by image search from the corpus using the ANN-generated image embeddings. The last three columns show the top-three images selected from the corpus using the SNN-generated image embeddings. It is obvious that image retrieval results, either using ANN or SNN, are visually closer to the query images compared with the randomly-selected images from the corpus. Again, our SNN implemented on the Loihi chip demonstrates comparable performance with the ANN.

\begin{center}
\includegraphics[scale=1.2]{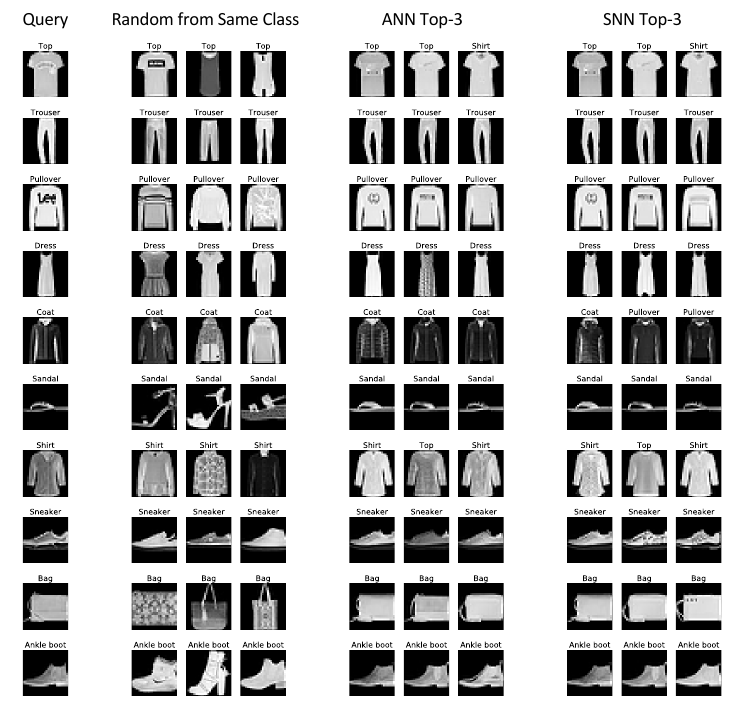}
\captionof{figure}[caption]{
\label{image_search}
Examples of the image retrieval by the artificial (ANN) and spiking (SNN) neural networks.
} 
\end{center}

The neural network inference latency (forward-pass runtime per example) comparison between the selected SNN and its ANN counterpart is shown in Table \ref{runtime_comparison}. Note that the Loihi could not support batch sizes larger than one at the time of the experiments. We can see that when the batch size equals one, the SNN on Loihi using 16 time steps has approximately 13.8x/11.3x longer runtime than the ANNs on Xeon/i7 CPUs, 3.8x longer than the ANN on ARM CPU, and 2.3x/2.5x longer than the ANNs on V100/T4 GPUs. The difference is even more dramatic if we use larger batch sizes for inference on the CPUs or GPUs. It is obvious that the SNNs on Loihi chip do not have an advantage in terms of the inference latencies. Several time steps that take an SNN to converge to its results leads to long execution times. Reducing the runtime is a direction where we look forward the neuromorphic hardware to improve upon.

\begin{table}
\centering
\caption{Inference latency.}
\begin{tabular}[t]{cccc}
\toprule
Model  Type & Batch Size & Hardware & Runtime  \\
& & & per Example (ms) \\
\midrule
ANN & 1 & Intel Xeon CPU (Gold 6148) & 0.216 \\
ANN & 262144 & Intel Xeon CPU (Gold 6148) & 0.0073 \\
ANN & 1 & Intel CPU (i7-8750H) & 0.2634 \\
ANN & 128 & Intel CPU (i7-8750H) & 0.013 \\
ANN & 1 & ARM CPU (Cortex-A72) & 0.778 \\
ANN & 2048 & ARM CPU (Cortex-A72) & 0.356 \\
ANN & 1 & NVIDIA GPU (V100) & 1.296 \\
ANN & 4096 & NVIDIA GPU (V100) & 0.0075 \\
ANN & 1 & NVIDIA GPU (T4) & 1.204 \\
ANN & 4096 & NVIDIA GPU (T4) & 0.010 \\
SNN (16) & 1 & Loihi chip (neuromorphic cores) & 2.984\\
SNN (128) & 1 & Loihi chip (neuromorphic cores) & 11.976\\
\bottomrule
\end{tabular}
\label{runtime_comparison}
\end{table}

The comparison of the average power consumptions between the SNNs and the ANNs is shown in Table \ref{power_comparison}. With the batch size set to one, the SNN with 16 time steps uses 217.0x/24.0x less power than the ANNs on Xeon/i7 CPUs, 9.3x less than the ANN on ARM CPU, and 40.8x/31.3x less than the ANNs on V100/T4 GPU. This is where neuromorphic hardware starts to shine as it consumes way less power than the conventional hardware. Utilizing the temporal sparsity of SNN appropriately, we believe the neuromorphic hardware can further reduce its power consumption. Another thing that we can observe from Table \ref{power_comparison} is that the static (idle) power dominates the power consumption of the Loihi chip.

\begin{table}
\centering
\caption{Inference power consumption.}
\begin{tabular}[t]{cccccc}
\toprule
Model & Batch & Hardware & Static & Dynamic & Total  \\
Type & Size & & Power (W) & Power (W) & Power (W) \\
\midrule
ANN & 1 & Intel Xeon CPU (Gold 6148) & 196 & 19.1 & 215.1 \\
ANN & 262144 & Intel Xeon CPU (Gold 6148) & 196 & 44.189 & 240.189 \\
ANN & 1 & Intel CPU (i7-8750H) & 22 & 1.805 & 23.805 \\
ANN & 128 & Intel CPU (i7-8750H) & 22 & 5.633 & 27.633 \\
ANN & 1 & ARM CPU (Cortex-A72) & 0.142 & 9.082 & 9.224 \\
ANN & 2048 & ARM CPU (Cortex-A72) & 0.142 & 2.698 & 2.84 \\
ANN & 1 & NVIDIA GPU (V100) & 24 & 16.441 & 40.441 \\
ANN & 4096 & NVIDIA GPU (V100) & 24 & 20.511 & 44.511 \\
ANN & 1 & NVIDIA GPU (T4) & 17 & 14.049 & 31.049 \\
ANN & 4096 & NVIDIA GPU (T4) & 17 & 18.228 & 35.228 \\
SNN (16) & 1 & Loihi chip (neuromorphic cores) & 0.946 & 0.044 & 0.991 \\
SNN (128) & 1 & Loihi chip (neuromorphic cores) & 0.952 & 0.064 & 1.016 \\
\bottomrule
\end{tabular}
\label{power_comparison}
\end{table}

We measured the total energy used per inference (forward pass) reported in Table \ref{energy_comparison}. These results can also be estimated by combining the results of Table \ref{runtime_comparison} and Table \ref{power_comparison}. As summarized in Table \ref{energy_comparison}, with the batch size set to one, the energy consumption of SNN with 16 time steps is 15.6x/3.2x less than the ANNs on Xeon/i7 CPUs, 2.5x less than the ANN on ARM CPU, and 17.5x/12.5x less than the ANNs on V100/T4 GPUs per inference. This proves the benefits of the neuromorphic hardware in the low energy-budget applications of machine learning, particularly lightweight image search engines and visual recommender systems. It is apparent that when large batch sizes are used, CPUs and GPUs consume less energy per example. However, there are many use cases where inference is executed in small batches, and they are the targets for neuromorphic hardware in the current stage.

\begin{table}
\centering
\caption{Inference energy consumption.}
\begin{tabular}[t]{ccccc}
\toprule
Model Type & Batch Size & Hardware & Energy per & Energy per\\
& & & Example (mJ) & Example (relative)\\
\midrule
ANN & 1 & Intel Xeon CPU (Gold 6148) & 46.787 & 15.6x \\
ANN & 262144 & Intel Xeon CPU (Gold 6148) & 1.753 & 0.585x \\
ANN & 1 & Intel CPU (i7-8750H) & 9.522 & 3.178x \\
ANN & 128 & Intel CPU (i7-8750H) & 0.316 & 0.105x \\
ANN & 1 & ARM CPU (Cortex-A72) & 7.379 & 2.463x \\
ANN & 2048 & ARM CPU (Cortex-A72) & 1.02 & 0.34x \\
ANN & 1 & NVIDIA GPU (V100) & 52.399 & 17.5x \\ 
ANN & 4096 & NVIDIA GPU (V100) & 0.337 & 0.112x \\ 
ANN & 1 & NVIDIA GPU (T4) & 37.399 & 12.5x \\ 
ANN & 4096 & NVIDIA GPU (T4) & 0.366 & 0.12x \\ 
SNN (16) & 1 & Loihi chip (neuromorphic cores) & 2.996 & 1x \\
SNN (128) & 1 & Loihi chip (neuromorphic cores) & 12.17 & 4.0x \\
\bottomrule
\end{tabular}
\label{energy_comparison}
\end{table}

Another observation is that the energy consumption for a small number of time steps does not scale linearly. For example, the energy consumption per inference for 128 time steps is only 4.0 times larger than 16 time steps (Table \ref{energy_comparison}). This is due to the constant portion of the energy needed for running each inference, which does not change by the number of time steps.

We use energy probes provided by Loihi NxSDK to perform the power and energy measurements of the Loihi chips. For the CPUs, we use Intelligent Platform Management Interface (IPMI) and the system profiler information to measure the power consumption and then, we integrate the power readings over time to get the energy consumption. For the GPUs, we use NVIDIA System Management Interface (nvidia-smi)\cite{nvidia2021smi} to measure the power consumption and again, we integrate the power readings to get the energy consumption.
\end{linenumbers}

\section*{Discussion}
\begin{linenumbers}
Our results confirm the energy efficiency of the Loihi neuromorphic chip. However, we noticed that the inference latency becomes impractically large when a network of Loihi chips are used. We ponder this is due to the interchip communication latencies. Nowadays in many applications, deep neural networks models with millions of parameters and billions of intermediate activations are used. Neuromorphic chips need to scale up, possibly by increasing the number of neuromorphic cores and on-chip memory, to support these applications in future.

The energy efficiency obtained by the Loihi chip in our experiments is owing to two factors. First, the model parameters are stored in the local memory of the neuromorphic cores, minimizing the energy cost of the data transfer to the shared memory. Second, the neuromorphic cores are optimized for specialized functionalities; this efficiency is very similar to that of other specialized accelerators e.g., graphical processing units (GPUs). The typical ANN-to-SNN conversion methods, including Algorithm \ref{alg} used here, do not capitalize on temporal sparsity, possible on the neuromorphic processors, as in the brain. So, designing better training and conversion algorithms to employ temporally sparse signals for neuromorphic machine learning is a promising future direction.

Finally, it is worthwhile to emphasize that to implement the complete image retrieval pipeline, we performed the nearest neighbor search on the host CPU cores. It is possible to carry out an approximate $k$-nearest neighbors algorithm on the neuromorphic chips\cite{frady2020neuromorphic}, but we believe that the CPU cores are largely needed for some stages of a machine learning pipeline. Thus, the role of neuromorphic computing is to improve the performance of special tasks and supplement the general-purpose processors.
\end{linenumbers}

\clearpage
\section*{Conclusion}
\begin{linenumbers}
We studied the application of the Loihi chip, a neuromorphic computing hardware developed by Intel, in image retrieval. Our results show that the generation of the deep learning embeddings by spiking neural networks for lightweight convolutional neural networks is about 2.5 times more energy-efficient compared with a CPU and 12.5 times more energy-efficient compared with a GPU. We confirm the long-term potential of neuromorphic computing in machine learning, not as a replacement for the predominant von Neumann architecture, but as accelerated coprocessors.
\end{linenumbers}

\section*{Acknowledgements}
\begin{linenumbers}
We would like to thank Hari Govind, Ramesh Subramonian, and Charles Leu at Target and Andreas Wild at Intel for helpful suggestions and constructive comments. We are also grateful of the Intel Neuromorphic Research Community for giving us access to the Loihi chips.
\end{linenumbers}

\section*{References}
\bibliography{neuromorphic}

\end{document}